\documentclass[conference]{IEEEtran}
\IEEEoverridecommandlockouts

\usepackage{cite}
\usepackage{amsmath,amssymb,amsfonts}
\usepackage{algorithmic}
\usepackage{graphicx}
\usepackage{textcomp}
\usepackage{xcolor}

\def\BibTeX{{\rm B\kern-.05em{\sc i\kern-.025em b}\kern-.08em
    T\kern-.1667em\lower.7ex\hbox{E}\kern-.125emX}}
\usepackage[utf8]{inputenc}
\usepackage{multirow}
\usepackage{booktabs}
\usepackage[compatibility=false]{caption}
\usepackage{url}
\usepackage{stfloats}
\usepackage{float}
\usepackage{microtype}
\usepackage[hidelinks]{hyperref}
\title{Towards Explainable  Sleep Stage Classification using Single-Channel Electroencephalogram}

\author{
\IEEEauthorblockN{Gurinder Kaur Sandhu$^{1}$, Jude Koenig$^{2}$, Ankita Shukla$^{2}$, Shruti Kshirsagar$^{1}$}
\IEEEauthorblockA{$^{1}$Wichita State University, Wichita, KS, USA}
\IEEEauthorblockA{$^{2}$University of Nevada, Reno, NV, USA}
\IEEEauthorblockA{gksandhu@shockers.wichita.edu}
}
\begin{document}

\maketitle

\begin{abstract}
Automated sleep staging has achieved strong benchmark 
accuracy; however, clinical adoption remains limited due to a lack of interpretability. 
In this work, we fill the research gap by extending the SleepXAI sleep 
staging framework to systematically compare five 
different XAI methods. Further, attribution fidelity is evaluated using Spearman rank 
correlation with AASM-defined EEG frequency bands, 
serving as clinically grounded reference signals. In this paper, we focus on  
single-channel Electroencephalogram (EEG), which is
more amenable to practical clinical deployment 
compared to full polysomnography setups.
Experiments are conducted on the Sleep-EDF benchmark 
dataset and the DREAMT dataset. In DREAMT, we 
stratified subjects based on OSA into four Apnea-Hypopnea Index (AHI) severity groups. We observed 
that the SleepXAI model achieves strong 
classification performance on both datasets. Among 
the evaluated methods, Integrated Gradients (IG) consistently achieves the highest mean attribution 
fidelity across all five sleep stages on Sleep-EDF 
and across all AHI groups on DREAMT, outperforming 
all other evaluated methods under the Spearman rank correlation metric. These findings highlight 
the importance of attribution method selection as a 
key factor for ensuring reliable and clinically 
interpretable automated sleep staging systems.
\end{abstract}



\section{Introduction}

Clinical AI deployment raises a question accuracy alone cannot answer: can a clinician trust a decision they cannot inspect~\cite{arrieta2020explainable}? Regulatory frameworks (FDA AI/ML action plan, EU AI Act) require diagnostic AI to demonstrate transparency before clinical adoption~\cite{muller2021trustworthy}. XAI methods address this by attributing predictions to specific input features, giving clinicians auditable evidence that decisions reflect physiological markers rather than spurious correlations~\cite{topol2019high}. Sleep staging is a key application of trustworthy clinical AI. Polysomnography (PSG) records multiple physiological signals during sleep. Trained technicians manually score each 30-second EEG epoch into one of five AASM-defined sleep stages: Wake, N1, N2, N3, and REM~\cite{berry2012aasm}. However, this manual process exhibits 15-20\% interscorer variability~\cite{Younes_Raneri_Hanly_2016a}.
Deep learning models now exceed 85\% accuracy on 
healthy-subjects benchmarks~\cite{eldele2021attention}, yet clinical 
adoption remains limited because these models provide 
no mechanism to verify whether predictions are 
grounded in AASM-defined EEG biomarkers: alpha 
(8-13~Hz), theta (4-7~Hz), spindles (12-15~Hz), 
delta (0.5-4~Hz), and sawtooth waves 
(2-6~Hz)~\cite{berry2012aasm}.

The challenge is compounded in Obstructive Sleep Apnea 
(OSA), where repeated apnea events introduce artifacts 
that obscure the spindles and delta waves staging 
models rely upon~\cite{korkal2021sleepfragmentation}. 
The Apnea-Hypopnea Index (AHI) stratifies OSA into 
four severity groups, yet the effect of AHI severity 
on both classification accuracy and attribution 
quality has received no systematic 
investigation~\cite{nassehi2024machine}. Without 
interpretable, biomarker-aligned explanations, it is 
impossible to determine whether a model has genuinely 
generalized to OSA-altered EEG or is exploiting 
fragmentation artifacts: a distinction that is 
both clinically meaningful and safety-critical. SleepXAI~\cite{dutt2023sleepxai} employs modified Grad-CAM as its sole attribution method, yet 
alternatives: Integrated Gradients 
(IG)~\cite{sundararajan2017axiomatic}, 
DeepLIFT~\cite{shrikumar2017learning}, SHAP 
GradientExplainer~\cite{lundberg2017unified}, and 
Vanilla Saliency~\cite{simonyan2013deep}  offer 
stronger axiomatic properties for 1D EEG signals. 
No prior study has compared these methods 
quantitatively across AHI severity groups, leaving 
a critical gap in the evidence base for trustworthy 
clinical sleep staging.

This paper closes both gaps by benchmarking five attribution methods on the Sleep-EDF dataset and the DREAMT dataset~\cite{wang2025dreamt} of 100 OSA patients stratified into four AHI groups.
We evaluate on the F4-M1 frontal channel, which is recommended by AASM guidelines for clinical sleep recording~\cite{berry2012aasm} and captures discriminative slow-wave and spindle activity relevant to sleep staging.
F4-M1 is anatomically proximate to the Fpz-Cz channel used in SleepXAI~\cite{dutt2023sleepxai}, enabling consistent comparison across both datasets.
Single-channel EEG also provides a controlled setting for evaluating XAI method sensitivity without the confounding effects of multi-channel interactions~\cite{nassehi2024machine, lv2024ssleepnet}. 
We use the same CNN-CRF architecture as SleepXAI~\cite{dutt2023sleepxai}, reimplemented in PyTorch instead of the original TensorFlow/Keras, and evaluate attribution fidelity via Spearman rank correlation with AASM frequency bands as clinically grounded ground truth.
The main contributions of this paper are

\begin{enumerate}
  \item We provide a systematic comparison of five XAI methods: modified Grad-CAM, Vanilla Saliency, DeepLIFT, SHAP, and IG. We evaluate them using AASM-defined frequency bands as a proxy for physiological alignment on the Sleep-EDF dataset, giving a detailed quantitative analysis for sleep stage classification.
  \item We present the first XAI evaluation based on different levels of OSA severity. We test and compare five methods particularly gradient based across four severity groups using the DREAMT dataset, which includes patients with different clinical conditions.
\end{enumerate}


The remainder of this paper is organized as follows. 
Section~\ref{sec:related} reviews related work. 
Section~\ref{sec:methodology} describes datasets, 
the CNN-CRF architecture, attribution methods, and 
evaluation framework and experimental setup. 
Section~\ref{sec:results} presents and discusses 
results. Section~\ref{limitations} discusses limitations. Section~\ref{sec:conclusion} concludes.
\section{Related Work}
\label{sec:related}
\subsection{Deep Learning for Sleep Staging}

Deep learning has been widely used across multiple domains~\cite{kshirsagar2022affective}. In the context of sleep analysis, 
automated sleep staging has evolved from handcrafted spectral pipelines~\cite{lajnef2015learning, tallal2026stda} to end-to-end architectures including DeepSleepNet~\cite{supratak2017deepsleepnet}, TinySleepNet~\cite{supratak2020tinysleepnet}, IITNet~\cite{back2019iitnet}, and Transformer-based models~\cite{eldele2021attention, phan2022xsleepnet}. 
Despite strong benchmark performance, deep learning sleep staging models
operate as black-box systems whose internal decisions are not directly
interpretable~\cite{sharma2023transparency, kshirsagartrustworthy}. This work focuses on evaluating XAI methods applied to the SleepXAI  architecture~\cite{dutt2023sleepxai} across Normal Subjects (SleepEDF Dataset) and OSA severity cohorts (Dreamt Dataset), rather than proposing a new architecture, it focuses on how explainable AI methods work on Normal subjects as well as AHI stratified OSA cohorts.

\subsection{Explainable AI for Sleep Staging}

Recent reviews have highlighted the growing need for explainability in deep learning-based sleep staging, particularly for clinical deployment~\cite{chen2024xaisleep, eldele2024attnEEG}. 
Among attribution methods applied to sleep EEG, Grad-CAM~\cite{selvaraju2017grad, dutt2023sleepxai, sharma2023transparency} has been most widely adopted. 
Ravindran and ContrerasVidal~\cite{sujatha2023empirical} found empirically that
path-integrated methods such as IG outperform activation-based methods for
temporal EEG localization. Lauritsen et al.~\cite{lauritsen2020explainable}
showed that DeepLIFT and IG outperform simpler gradient methods for clinical
event prediction. The Spearman rank correlation with the AASM-defined EEG frequency
bands has been proposed as a principled quantitative evaluation
framework~\cite{sujatha2023empirical,samek2016evaluating}.

\subsection{Sleep Staging in Patients with OSA}

Most sleep staging benchmarks use healthy adult recordings. OSA patients
exhibit disrupted sleep architecture causing high-frequency noise bursts that
obscure spindles and delta waves critical for N2 and N3
staging~\cite{korkal2021sleepfragmentation}. SSleepNet~\cite{lv2024ssleepnet}
examined how apnea severity affects staging accuracy on the SHHS dataset,
finding monotonic accuracy degradation, but did not apply XAI methods. Moreover, Hossain and Kshirsagar\cite{hossain2026demographic} showed that stratified fine-tuning along with patient demographics improved sleep staging performance. Furthermore, Tallal et al.~\cite{tallal2026modulation} demonstrated that
modulation spectral features improve sleep stage classification
across apnea-based cohorts, yet interpretability of model decisions across
AHI severity groups remained unaddressed.

\section{Methodology}
\label{sec:methodology}
\begin{figure}[t]
  \centering
  \includegraphics[width=\columnwidth]{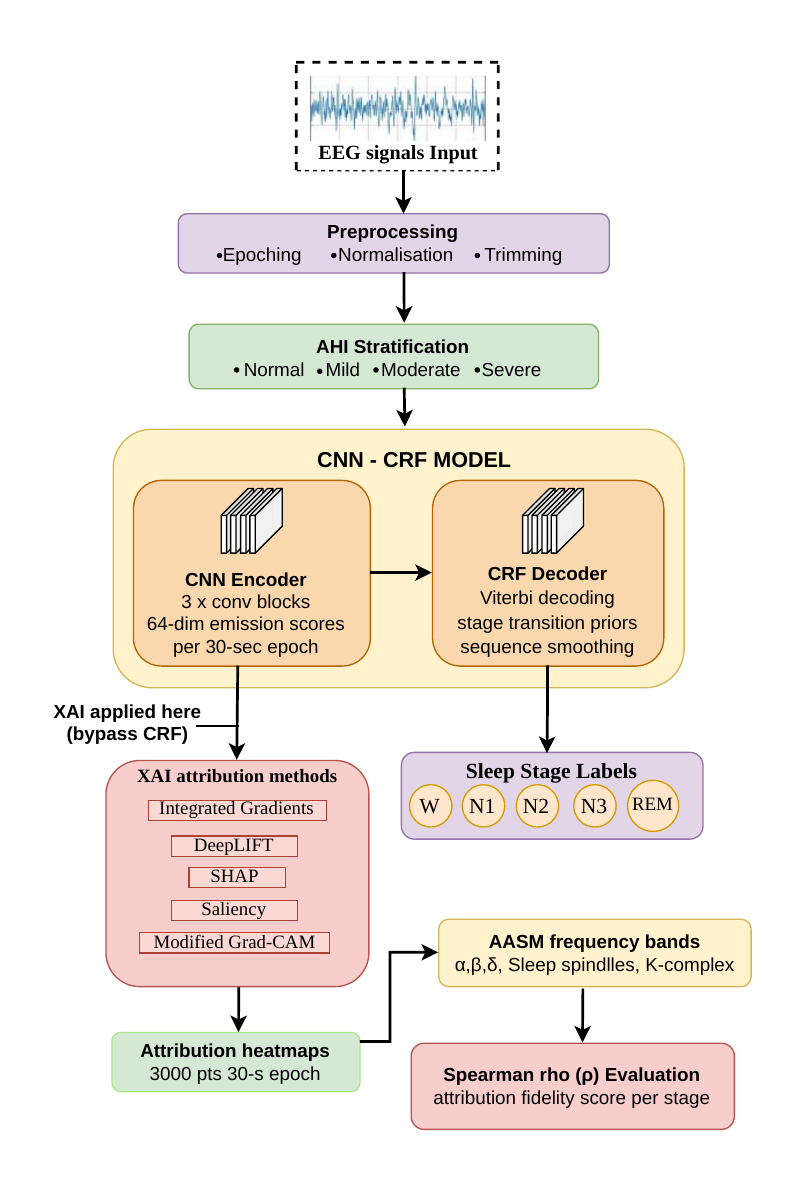}
  \caption{End-to-end pipeline of the experimental framework adopted from SleepXAI~\cite{dutt2023sleepxai}. The CNN-CRF architecture outperformed the CNN-CNN baseline and was selected for XAI evaluation.}
  \label{fig:results_p3}
\end{figure}

This section describes the datasets, preprocessing 
pipeline, CNN-CRF model architecture, XAI attribution 
methods, and quantitative evaluation framework used 
in this study. Fig. \ref{fig:results_p3} illustrates the overall experimental pipeline. Raw EEG signals are preprocessed and stratified into four AHI severity groups before being classified using the CNN-CRF model. The CNN-CRF model was selected over the CNN-CNN baseline following performance comparison on Sleep-EDF, as detailed in Section~\ref{sec:results}. Five XAI methods are then applied to the CNN encoder to generate attribution heatmaps, which are evaluated against AASM-defined EEG frequency bands using Spearman rank correlation.

\subsection{Datasets}
For this work, we employed two databases described below:
\subsubsection{SleepEDF}
The Sleep Cassette dataset \cite{kemp2000sleepedf} includes 39 recordings from 20 healthy adults aged 25 to 34. The data was recorded at a sampling rate of 100 Hz and labeled using the AASM five-stage sleep scoring system. We removed segments with body movement and kept only the data within 30 minutes before the first sleep stage and 30 minutes after the last sleep stage. After this cleaning, we kept 42,308 labeled segments. We selected and analyzed single EEG channel: Fpz - Cz (front to center of the head) which gave best accuracy in SleepXAI. we followed the same preprocessing steps like SleepXAI~\cite{dutt2023sleepxai} to evaluate the performance of the model.

\subsubsection{DREAMT}
The DREAMT dataset was collected from 100 adult patients with OSA at Duke University~\cite{wang2025dreamt}. It includes six EEG channels recorded at 100~Hz and labeled by experts using AASM guidelines. The average age of the participants was $56.2 \pm 16.6$ years. The average body mass index (BMI) was $33.7 \pm 8.6$~kg/m$^2$, and the average apnea-hypopnea index (AHI) was $22.1 \pm 28.7$~events/h. In this study, we used single channel: \textit{F4 - M1} (a frontal channel similar to Fpz - Cz in SleepEDF). We divided the patients into four groups based on AHI: Normal ($<5$), Mild (5--14), Moderate (15--29), and Severe ($\geq 30$~events/h)~\cite{berry2012aasm, tallal2026modulation}. Table~\ref{tab:ahi} shows the number of data segments in each group. Sleep stages N1 and N3 appear less often in all groups, which makes them harder to analyze.

\begin{table}[t]
\centering
\caption{AHI Based Stratification of DREAMT Dataset.
         N1 and N3 Are Consistently Underrepresented
         Across All OSA Severity Groups.}
\label{tab:ahi}
\small
\setlength{\tabcolsep}{3.5pt}
\begin{tabular}{lcccccccc}
\toprule
{Group} & {AHI} & {N} & 
{W} & {N1} & {N2} & 
{N3} & {REM} & {Total}\\
\midrule
Normal   & $<$5 & 26 & 4847 & 1715 & 10236 
         & 567 & 2506 & 19871 \\
Mild     & 5-14 & 25 & 4164 & 1967 & 10286 
         & 906 & 2291 & 19614 \\
Moderate & 15-29 & 24 & 4517 & 2185 &  9805 
         & 647 & 1905 & 19059 \\
Severe   & $\geq$30 & 25 & 4628 & 2949 &  9593 
         & 584  & 1674 & 19419 \\
\bottomrule
\end{tabular}
\end{table}

\subsection{Data Preprocessing}

The DREAMT~\cite{wang2025dreamt} preprocessing pipeline follows six steps, consistent with SleepXAI~\cite{dutt2023sleepxai}. 
First, ambiguous P-stage epochs are removed. Second, a single EEG channel F4-M1 is selected where M1 denotes the left mastoid reference electrode. 
Third, sleep stage labels are mapped to numerical values. Fourth, EEG signals are segmented into non-overlapping 30-second epochs. 
At a sampling rate of 100~Hz, each epoch contains 3,000 samples: this epoch length is the AASM standard for manual sleep scoring~\cite{berry2012aasm}. 
Fifth, to reduce class imbalance caused by extended wake periods at the recording start and end, only epochs within $\pm$30 minutes of the first and last non-wake epochs are retained~\cite{dutt2023sleepxai}. 
Finally, signals are normalized using wake-stage statistics:
\begin{equation}
  \hat{x} = \frac{x - \mu_x}{\sigma_{\text{wake}}},
  \label{eq:norm}
\end{equation}
where $x$ denotes the original EEG signal (time series) acquired from a single subject before normalization, $\hat{x}$ represents the normalized EEG signal after applying the normalization procedure, $\mu_x$ denotes the mean of the entire EEG recording for that subject, and $\sigma_{\mathrm{wake}}$ represents the standard deviation calculated exclusively from the wake-stage EEG epochs of the same subject. The same pipeline is applied to Sleep-EDF, with Fpz-Cz channels used instead of F4-M1. We apply Leave-One-Subject-Out (LOSO) cross-validation for both datasets, following the SleepXAI protocol~\cite{dutt2023sleepxai}. A single CNN-CRF model is trained per AHI group on the DREAMT dataset; no subject's data appears in both training and evaluation.

\subsection{Model Architecture}

We used the same CNN-CRF architecture as SleepXAI~\cite{dutt2023sleepxai}. The model consists of two main components.

\begin{enumerate}
  \item Sleep Encoder: This module converts each 
  3,000-sample epoch into a 64-dimensional feature 
  representation. It uses three convolutional blocks 
  with 128, 64, and 32 filters, respectively, each with 
  a kernel size of 10 and SELU activation. Each block 
  is followed by max pooling with window size 2 and stride 2, and dropout for 
  regularization. The network parameters are initialized 
  using Glorot initialization~\cite{glorot2010understanding}.

  \item SleepLabeler with CRF Decoder: The sequence 
  of epoch embeddings (length 100) is then passed through 
  two Conv1D layers with 128 filters and kernel size 3. 
  These layers produce emission scores for each epoch. 
  A linear chain conditional random field (CRF)~\cite{lafferty2001conditional} 
  is used to decode the final sleep stages using the Viterbi algorithm.
\end{enumerate} The CRF learns transition probabilities between sleep 
stages, allowing it to enforce realistic physiological 
patterns (e.g., direct transitions from N3 to Wake are 
unlikely). This helps correct errors caused by noisy or 
ambiguous epochs, especially in OSA conditions.


\subsection{Explainable AI Methods}

All five XAI methods are applied to the predicted class 
score at the CNN encoder output, computed before the CRF 
decoder. Each method produces a 3,000-point attribution heatmap 
aligned with the raw EEG input. This follows the same approach as SleepXAI~\cite{dutt2023sleepxai}, which also applied Grad-CAM to the CNN encoder. This design choice is justified on two grounds. First, the CNN encoder is the sole component responsible for feature learning from raw EEG; the CRF decoder performs sequence-level smoothing using the CNN's emission scores as input, but does not itself learn EEG features. Second, the emission scores produced by the CNN are the direct and only input to the CRF, so attributions computed on the emission scores indirectly explain the information that drives the final CRF prediction. Applying XAI methods to the CRF output would require CRF-specific attribution techniques that are not yet standardized for clinical EEG analysis~\cite{arrieta2020explainable, Sousa2023}.

\subsubsection{Modified GradCAM}
Modified GradCAM~\cite{selvaraju2017grad, dutt2023sleepxai} 
adapts the original 2D GradCAM formulation to 1D EEG 
signals. Gradients are computed with respect to the final 
convolutional layer~\cite{selvaraju2017grad, dutt2023sleepxai} (output size $32\!\times\!718$), and 
the resulting activation map is upsampled to 3,000 points.

\subsubsection{Vanilla Saliency}
Vanilla Saliency~\cite{simonyan2013deep} computes the 
absolute gradient of the predicted class score with 
respect to the raw input, without considering any 
reference or baseline state of the network.

\subsubsection{DeepLIFT}
DeepLIFT~\cite{shrikumar2017learning} attributes prediction differences to input features relative to a reference baseline. We use the Captum DeepLIFT implementation with a zero baseline $x' = \mathbf{0}$, which in Captum approximates the rescale rule as a gradient$\times$input form:
\begin{equation}
  \text{attr}_i = (x_i - x'_i) \times 
  \tfrac{g_{x_i} + g_{x'_i}}{2},
  \label{eq:deeplift}
\end{equation}
where $x_i$ is the $i$-th input feature, $x'_i$ is the corresponding baseline value (zero), and $g_{x_i}$, $g_{x'_i}$ denote the gradients of the output with respect to $x_i$ evaluated at the input and baseline respectively. This approximation helps reduce gradient saturation effects 
commonly observed in vanilla saliency methods. The zero 
baseline ($x' = \mathbf{0}$) serves as a neutral reference 
point~\cite{ancona2018towards}, though after wake-statistics normalization it corresponds to the subject's mean EEG level rather than true signal absence.

\subsubsection{SHAP GradientExplainer}    

SHAP GradientExplainer~\cite{lundberg2017unified}
computes expected gradients by averaging gradient$\times$input 
attributions over 20 background EEG epochs sampled from 
the same subject, providing a data-driven reference 
distribution rather than a fixed zero baseline.

\subsubsection{Integrated Gradients (IG)}
IG~\cite{sundararajan2017axiomatic} 
computes attributions by integrating gradients along a 
straight-line path from a zero baseline $x' = \mathbf{0}$ to the input over $m\!=\!50$ 
steps:
\begin{equation}
  \text{IG}_i(x) = (x_i - x'_i) \times 
  \frac{1}{m}\sum_{k=1}^{m}
  \frac{\partial F\!\left(x' + 
  \frac{k}{m}(x - x')\right)}{\partial x_i},
  \label{eq:ig}
\end{equation}
where $x_i$ is the $i$-th input feature, $x'_i$ is the corresponding zero baseline, $m$ is the number of interpolation steps, where $F$ denotes the CNN prediction function that maps an input EEG epoch to the corresponding sleep-stage prediction. The zero baseline is a standard choice for EEG signals~\cite{ancona2018towards}, as a flat signal represents the absence of neural activity and provides a physiologically neutral reference point. IG satisfies the completeness 
property, ensuring that the attributions sum to the 
total prediction difference, $\sum_i \text{IG}_i(x) = F(x) - F(x')$, enabling consistent 
and interpretable explanations across the full 
30-second epoch. At $m\!=\!50$ steps, the approximation error (completeness gap, $\delta = |\sum_i \text{IG}_i(x) - (F(x)-F(x'))|$) is small in practice; this step count is consistent with standard IG implementations~\cite{sundararajan2017axiomatic}.
\subsection{Evaluation Framework}

We evaluate model performance along two 
complementary axes: classification accuracy 
on the sleep staging task, and attribution 
fidelity of the five XAI methods against 
AASM-defined frequency bands as one proxy for physiological alignment.

\subsubsection{Accuracy}
It measures the 
proportion of correctly classified epochs 
across all five sleep stages:
\begin{equation}
\text{Acc} = \frac{\sum_{c=1}^{5} \text{TP}_c}
{N} \times 100
\end{equation}
where $N$ is the total number of epochs. 
\subsubsection{PerStage F1Score} It is the harmonic 
mean of precision and recall for each stage 
$c \in \{\text{W, N1, N2, N3, REM}\}$:
\begin{equation}
\text{F1}_c = \frac{2 \cdot \text{TP}_c}
{2 \cdot \text{TP}_c + \text{FP}_c + \text{FN}_c}
\end{equation}
where $\text{TP}_c$, $\text{FP}_c$, and 
$\text{FN}_c$ denote true positives, false 
positives, and false negatives for stage $c$, 
respectively. 
Perstage F1 is the primary 
metric for evaluating minority classes N1 
and N3, whose subtle EEG signatures make 
them\textbf{} consistently the hardest stages to 
classify~\cite{dutt2023sleepxai, zhao2022deep}.

\subsubsection{Spearman Rank Correlation (\texorpdfstring{$\rho$}{rho})}
We use Spearman rank correlation to quantify the alignment between each XAI method's at\textbf{}tribution heatmap and the instantaneous amplitude envelope of the AASM-defined dominant EEG frequency band for each sleep stage~\cite{sujatha2023empirical}:
\begin{equation}
\rho = 1 - \frac{6 \sum_{k=1}^{n} d_k^2}
{n(n^2 - 1)}
\end{equation}
where $d_k = \text{rank}(A_k) - \text{rank}(G_k)$, $A_k$ is the attribution intensity at time point $k$, $G_k$ is the band-pass filtered EEG amplitude at the same point, and $n = 3{,}000$ samples per 30-second epoch.
Attributions are used in their signed form before computing the correlation; negative values indicate time points where the attribution decreases relative to the zero baseline.
For statistical comparison, we compute one $\rho$ value per epoch; 95\% confidence intervals and significance tests are then computed over the distribution of per-epoch $\rho$ values across all epochs of a given stage, treating epochs as the unit of analysis.
This metric is appropriate for three reasons. First, AASM guidelines define sleep stages by their dominant EEG frequency bands, specifically delta (0.5--4~Hz) for N3, spindles (12--15~Hz) for N2, and alpha (8--13~Hz) for Wake, providing a clinically validated, expert-defined reference signal against which attribution quality can be objectively measured~\cite{berry2012aasm} (Table~\ref{tab:aasm_bands}). 

\begin{table}[h]
\centering
\caption{AASM-Defined EEG Frequency Bands Used as Ground Truth}
\label{tab:aasm_bands}
\small
\resizebox{\columnwidth}{!}{%
\begin{tabular}{lccl}
\toprule
Sleep Stage & Band (Symbol) & Range (Hz) & EEG Feature \\
\midrule
Wake & Alpha ($\alpha$) & 8--13 Hz & Alpha rhythm \\
N1   & Theta ($\theta$) & 4--7 Hz  & Theta waves \\
N2   & Sigma ($\sigma$) & 12--15 Hz & Sleep spindles \\
N3   & Delta ($\delta$) & 0.5--4 Hz & Slow/delta waves \\
REM  & Mixed (--)        & 2--6 Hz  & Sawtooth waves \\
\bottomrule
\end{tabular}%
}
\end{table}



Table~\ref{tab:aasm_bands} summarizes the AASM-defined frequency band and characteristic EEG feature used as ground truth for each sleep stage~\cite{berry2012aasm}.The source code is publicly available at \url{https://github.com/GeetSandhu/xai-2d--eeg-sleep-staging}.

\begin{table*}[t]
\centering
\caption{Classification Results with single-channel EEG signals on the SleepEDF and DREAMT dataset}
\label{tab:dreamt_full}
\small
\begin{tabular}{lllccccccc}
\hline
{Dataset} &{Group} & {Model} & {Channel} & {Acc (\%)} & {W } & {N1} & {N2} & 
{N3} & {REM}\\
\hline
\multirow{2}{*}{SleepEDF} & \multirow{2}{*}{Ref~\cite{dutt2023sleepxai}}& CNN - CNN & Fpz - Cz & 84.79 & 90.95 & 48.39 & 87.28 & 84.83 & 85.56\\
 &  & CNN - CRF & Fpz - Cz & \textbf{86.81} & \textbf{91.84} & \textbf{62.99} & \textbf{88.17} & \textbf{86.85} & \textbf{86.88}\\  
\hline
\multirow{2}{*} {DREAMT} & \multirow{2}{*}{Normal}
& CNN - CNN & F4 - M1 & 74.73 & 80.32 & 27.66 & 84.00 & 30.55 & 60.64\\
& & CNN - CRF & F4 - M1 & \textbf{76.10} & \textbf{82.92} & \textbf{32.90} & \textbf{84.65} & \textbf{33.18} & \textbf{64.36}\\
\hline

\multirow{2}{*}{DREAMT} & \multirow{2}{*}{Mild}
& CNN - CNN & F4 - M1 & 66.00 & 69.67 & 21.88 & 77.42 & 27.10 & 54.63\\
&& CNN - CRF & F4 - M1 & \textbf{70.02} & \textbf{71.54} & 27.46 & 82.20 & 38.51 & \textbf{54.65}\\
\hline

\multirow{2}{*}{DREAMT} & \multirow{2}{*}{Moderate}
& CNN - CNN & F4 - M1 & 70.09 & 77.49 & 26.61 & \textbf{83.27} & 27.01 & 60.14\\
&& CNN - CRF & F4 - M1 & \textbf{73.02} & \textbf{79.10} & \textbf{30.72} & 83.06 & \textbf{35.09} & \textbf{61.62}\\
\hline

\multirow{2}{*}{DREAMT} & \multirow{2}{*}{Severe}
& CNN - CNN & F4 - M1 & 63.32 & 68.39 & 23.99 & 77.58 & 32.02 & 32.54\\
&& CNN - CRF & F4 - M1 & \textbf{67.49} & \textbf{70.54} & \textbf{27.23} & \textbf{79.51} & \textbf{47.76} & \textbf{56.52}\\
\hline

\end{tabular}%
\end{table*}

\section{Results and Discussion}
\label{sec:results}
This section presents classification and XAI 
attribution results across both datasets. Findings 
are discussed in the context of existing literature.

\subsection{Sleep Staging Performance Across OSA Severity Groups}

We evaluated the robustness of the SleepXAI (CNN-CRF) model across different OSA severity levels of DREAMT dataset. We compared its performance with the baseline CNN-CNN model using single EEG channel (F4-M1). We also report results on the Sleep-EDF dataset~\cite{dutt2023sleepxai} to validate our implementation and provide a benchmark for comparison.
Table~\ref{tab:dreamt_full} summarizes the classification accuracy and per-stage F1-scores. On the Sleep-EDF dataset, the CNN-CRF model consistently outperforms the CNN-CNN model. We then train a separate CNN-CRF model on each DREAMT AHI group using the same architecture and hyperparameters, 
following the LOSO cross-validation protocol~\cite{dutt2023sleepxai} to evaluate sleep staging performance across four AHI severity groups.

The CNN-CRF model consistently outperforms the CNN-CNN model across all AHI groups on the DREAMT dataset. The CRF decoder models temporal dependencies between consecutive sleep stages and reduces isolated prediction errors. The F4-M1 frontal channel consistently captures discriminative slow-wave and spindle activity across all AHI severity groups, consistent with the frontal EEG advantage reported in prior sleep staging studies~\cite{korkal2021sleepfragmentation}. Classification accuracy decreases as OSA severity increases. The Normal group achieves the highest accuracy, whereas the Severe group achieves the lowest. OSA introduces frequent arousals and sleep fragmentation that distort EEG patterns and make sleep stages harder to classify~\cite{tallal2026modulation}. Similar trends have been reported by SleepNet on the SHHS dataset~\cite{lv2024ssleepnet}. Compared with Sleep-EDF, the DREAMT dataset shows 8-10\% reduction in accuracy. This reduction likely reflects the greater variability in clinical OSA recordings, though differences in participant demographics (age, BMI, sex), recording device, and EEG channel between datasets are potential confounders that were not formally tested~\cite{arrieta2020explainable}. Interestingly, the Moderate group achieves slightly higher accuracy than the Mild group on F4-M1 channel. This finding suggests that the relationship between OSA severity and sleep staging performance may not be strictly monotonic. Larger stratified cohorts are needed to determine whether this trend is consistent across broader clinical populations~\cite{nassehi2024machine}. The N1 stage consistently achieves the lowest F1-score across all AHI groups. N1 is a transitional sleep stage with subtle EEG characteristics and fewer training samples than the other stages. These factors make it the most difficult stage to classify under class imbalance~\cite{zhao2022deep,tallal2026modulation}.

\begin{table*}[t]
\centering
\caption{XAI Spearman Correlation ($\rho$) with AASM Ground Truth for Sleep-EDF Dataset (Fpz-Cz). Values shown as $\rho$ [95\% CI]. * = Bonferroni-corrected significance ($p < 0.002$). Bold = best per stage.}
\label{tab:xai_correlation}
\setlength{\tabcolsep}{4pt}
\footnotesize
\begin{tabular}{lccccc}
\toprule
Method & Wake & N1 & N2 & N3 & REM \\
\midrule
Grad-CAM & 0.020 [-0.052, 0.091]* & 0.009 [-0.067, 0.084]* & 0.025 [-0.046, 0.096]* & 0.015 [-0.056, 0.086]* & 0.002 [-0.071, 0.075] \\
Saliency & 0.035 [-0.035, 0.105]* & 0.068 [-0.003, 0.138]* & 0.131 [0.062, 0.200]* & 0.005 [-0.066, 0.076] & 0.121 [0.051, 0.190]* \\
DeepLIFT & 0.045 [-0.026, 0.115]* & 0.075 [0.004, 0.145]* & 0.138 [0.069, 0.206]* & 0.121 [0.051, 0.191]* & 0.139 [0.070, 0.208]* \\
SHAP & 0.027 [-0.043, 0.097]* & 0.042 [-0.029, 0.113]* & 0.091 [0.021, 0.161]* & 0.230 [0.161, 0.296]* & 0.084 [0.014, 0.154]* \\
\textbf{IG} & \textbf{0.092 [0.022, 0.162]*} & \textbf{0.118 [0.048, 0.188]*} & \textbf{0.204 [0.135, 0.270]*} & \textbf{0.290 [0.224, 0.354]*} & \textbf{0.199 [0.130, 0.265]*} \\
\bottomrule
\end{tabular}
\end{table*}

\begin{table*}[t]
\centering
\caption{XAI Spearman Correlation ($\rho$) with AASM Ground Truth for DREAMT Dataset. Values shown as $\rho$ [95\% CI]. * = Bonferroni-corrected significance ($p < 0.0005$). Bold = best per group.}
\label{tab:xai_correlation2}
\setlength{\tabcolsep}{4pt}
\footnotesize
\begin{tabular}{llccccc}
\toprule
Group & Method & Wake & N1 & N2 & N3 & REM \\
\midrule
\multirow{5}{*}{Normal}
 & Grad-CAM & 0.017 [-0.072, 0.105]* & -0.002 [-0.092, 0.088] & 0.013 [-0.076, 0.101]* & 0.024 [-0.144, 0.191]* & 0.004 [-0.096, 0.103] \\
 & Saliency  & 0.026 [-0.060, 0.112]* & 0.040 [-0.047, 0.126]* & 0.068 [-0.019, 0.153]* & -0.025 [-0.192, 0.143] & 0.046 [-0.047, 0.138]* \\
 & DeepLIFT  & 0.045 [-0.041, 0.131]* & 0.060 [-0.027, 0.146]* & 0.074 [-0.012, 0.159]* & 0.100 [-0.069, 0.263]* & 0.081 [-0.012, 0.172]* \\
 & SHAP      & 0.049 [-0.037, 0.135]* & 0.043 [-0.043, 0.129]* & 0.055 [-0.032, 0.140]* & 0.247 [0.083, 0.398]* & 0.055 [-0.038, 0.146]* \\
 & \textbf{IG} & \textbf{0.086 [-0.001, 0.171]*} & \textbf{0.084 [-0.003, 0.169]*} & \textbf{0.094 [0.008, 0.179]*} & \textbf{0.314 [0.155, 0.457]*} & \textbf{0.127 [0.034, 0.216]*} \\
\midrule
\multirow{5}{*}{Mild}
 & Grad-CAM & 0.014 [-0.081, 0.108]* & -0.000 [-0.096, 0.095] & 0.017 [-0.075, 0.108]* & 0.033 [-0.116, 0.180]* & -0.001 [-0.110, 0.109] \\
 & Saliency  & 0.035 [-0.056, 0.125]* & 0.032 [-0.058, 0.121]* & 0.060 [-0.029, 0.148]* & -0.090 [-0.235, 0.058]* & 0.035 [-0.062, 0.132]* \\
 & DeepLIFT  & 0.052 [-0.039, 0.142]* & 0.049 [-0.042, 0.138]* & 0.071 [-0.018, 0.158]* & 0.021 [-0.128, 0.169] & 0.061 [-0.037, 0.157]* \\
 & SHAP      & 0.042 [-0.049, 0.132]* & 0.029 [-0.061, 0.118]* & 0.049 [-0.040, 0.138]* & 0.216 [0.071, 0.353]* & 0.047 [-0.050, 0.144]* \\
 & \textbf{IG} & \textbf{0.090 [-0.001, 0.179]*} & \textbf{0.074 [-0.016, 0.163]*} & \textbf{0.091 [0.002, 0.178]*} & \textbf{0.197 [0.051, 0.335]*} & \textbf{0.108 [0.011, 0.203]*} \\
\midrule
\multirow{5}{*}{Moderate}
 & Grad-CAM & 0.020 [-0.074, 0.113]* & 0.004 [-0.089, 0.096] & 0.005 [-0.086, 0.095]* & 0.015 [-0.118, 0.148]* & 0.000 [-0.101, 0.102] \\
 & Saliency  & 0.018 [-0.072, 0.108]* & 0.030 [-0.060, 0.120]* & 0.033 [-0.057, 0.122]* & -0.087 [-0.218, 0.046]* & 0.053 [-0.044, 0.148]* \\
 & DeepLIFT  & 0.037 [-0.053, 0.127]* & 0.053 [-0.037, 0.142]* & 0.042 [-0.048, 0.132]* & 0.038 [-0.096, 0.170]* & 0.093 [-0.003, 0.188]* \\
 & SHAP      & 0.047 [-0.043, 0.136]* & 0.037 [-0.053, 0.127]* & 0.034 [-0.056, 0.124]* & 0.223 [0.092, 0.346]* & 0.061 [-0.036, 0.156]* \\
 & \textbf{IG} & \textbf{0.098 [0.009, 0.187]*} & \textbf{0.091 [0.001, 0.179]*} & \textbf{0.070 [-0.020, 0.159]*} & \textbf{0.268 [0.140, 0.387]*} & \textbf{0.137 [0.041, 0.230]*} \\
\midrule
\multirow{5}{*}{Severe}
 & Grad-CAM & 0.012 [-0.085, 0.109]* & 0.001 [-0.098, 0.100] & 0.013 [-0.083, 0.108]* & 0.015 [-0.148, 0.177]* & 0.002 [-0.109, 0.114] \\
 & Saliency  & 0.010 [-0.080, 0.099] & 0.029 [-0.061, 0.119]* & 0.052 [-0.038, 0.141]* & -0.032 [-0.194, 0.132]* & 0.068 [-0.028, 0.163]* \\
 & DeepLIFT  & 0.026 [-0.064, 0.115]* & 0.047 [-0.043, 0.137]* & 0.061 [-0.029, 0.150]* & 0.062 [-0.101, 0.223]* & 0.097 [0.001, 0.191]* \\
 & SHAP      & 0.036 [-0.054, 0.125]* & 0.037 [-0.053, 0.126]* & 0.044 [-0.046, 0.133]* & 0.201 [0.040, 0.352]* & 0.065 [-0.031, 0.160]* \\
 & \textbf{IG} & \textbf{0.078 [-0.012, 0.166]*} & \textbf{0.076 [-0.014, 0.165]*} & \textbf{0.088 [-0.002, 0.176]*} & \textbf{0.222 [0.062, 0.371]*} & \textbf{0.132 [0.036, 0.225]*} \\
\bottomrule
\end{tabular}
\end{table*}

\subsection{XAI Attribution Analysis}
We evaluated the clinical reliability of the five XAI methods by measuring how well their attribution maps align with established AASM-defined EEG biomarkers. We computed the Spearman rank correlation between the attribution heatmaps and the corresponding sleep-stage frequency bands on both the Sleep-EDF and DREAMT datasets.
Tables~\ref{tab:xai_correlation} and~\ref{tab:xai_correlation2}
report Spearman correlations with 95\% confidence intervals 
between attribution heatmaps and AASM-defined EEG frequency 
bands on Sleep-EDF and DREAMT respectively. Significance 
stars reflect Bonferroni-corrected p-values; with n=3,000 
samples per epoch, small effects can reach statistical 
significance even when the 95\% CI includes zero. The ranking 
IG $>$ SHAP $>$ DeepLIFT $>$ Saliency $>$ Grad-CAM is 
consistent across all settings, identifying IG as the most 
clinically aligned method for both healthy and OSA 
groups~\cite{sujatha2023empirical, arrieta2020explainable}.

IG achieves the highest correlation on Sleep-EDF, with N3 
$\rho=0.290$ [0.224, 0.354] and N2 $\rho=0.204$ [0.135, 0.270] 
both showing the strongest alignment among all evaluated methods~\cite{sujatha2023empirical}, and significant N1 
correlation ($\rho=0.118$ [0.048, 0.188]) reflecting clearer 
theta-band activity in healthy subjects compared to OSA cohorts 
where apnea events disrupt N1 theta waves. All five stages are 
significant ($p < 0.002$, Bonferroni-corrected across 
25 comparisons: 5 methods $\times$ 5 stages). Its path-based 
integration captures cumulative temporal contributions of EEG 
signals~\cite{sundararajan2017axiomatic}. N2 and N3 correlations 
are highest among all stages across AHI groups, indicating robust learning of 
spindle and slow-wave patterns despite OSA-related 
fragmentation~\cite{tallal2026modulation, korkal2021sleepfragmentation}. 
N3 consistently achieves the highest IG correlation across all 
cohorts because delta waves (0.5--4~Hz) are the most temporally 
prominent EEG feature, making them easier for IG to localize 
compared to the subtler alpha and theta patterns of Wake and N1 stages.

Grad-CAM performs worst on both datasets, with small positive correlations for N3 and REM on Sleep-EDF and the lowest mean across DREAMT groups. Its $32\!\times\!718$ activation map is upsampled to 3,000 points, which limits temporal precision independently of explanation quality and partially confounds its ranking against input-resolution methods~\cite{selvaraju2017grad, dutt2023sleepxai}. 
SHAP achieves the second-highest N3 correlation on Sleep-EDF 
($\rho=0.230$ [0.161, 0.296]), attributable to its 
physiologically realistic background 
distribution~\cite{lundberg2017unified}. DeepLIFT ranks third 
on Sleep-EDF by avoiding gradient 
saturation~\cite{shrikumar2017learning}, but its instability 
under OSA-specific EEG confirms that saturation avoidance alone 
is insufficient for robust alignment under this metric.

\begin{figure}
  \centering
  \includegraphics[width=\columnwidth]{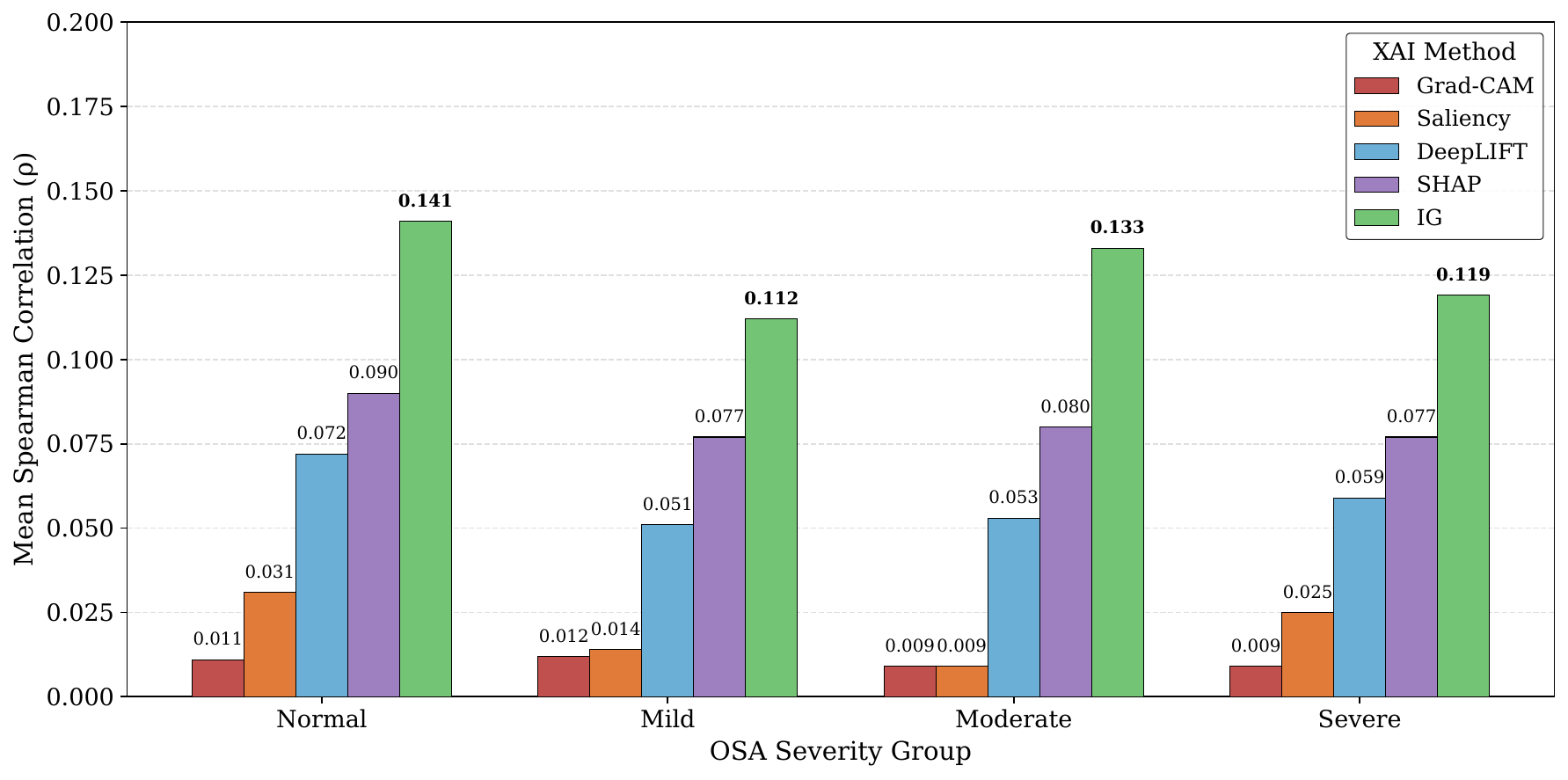}
  \caption{Mean Spearman correlation ($\rho$) between XAI attribution heatmaps and AASM-defined EEG frequency bands across four AHI severity groups (DREAMT F4-M1).}
  \label{fig:ig_heatmaps}
\end{figure}

Fig.~\ref{fig:ig_heatmaps} shows mean Spearman correlation 
across all five XAI methods and four AHI severity groups of DREAMT dataset. 
IG consistently achieves the highest attribution fidelity 
across all cohorts, with mean $\rho=0.141$ for Normal, 
$\rho=0.112$ for Mild, $\rho=0.133$ for Moderate, and 
$\rho=0.119$ for Severe. The ranking IG $>$ SHAP $>$ 
DeepLIFT $>$ Saliency $>$ Grad-CAM holds on average across all cohorts. Saliency and DeepLIFT produce near-zero correlations 
on Mild and Moderate cohorts, indicating unreliable 
physiological alignment under moderate apnea burden. 
The non-monotonic pattern where Moderate outperforms Mild 
is consistent across most methods and aligns with the 
accuracy trends in Table~\ref{tab:dreamt_full}.

\begin{figure}[!ht]
  \centering
  \includegraphics[width=\columnwidth]{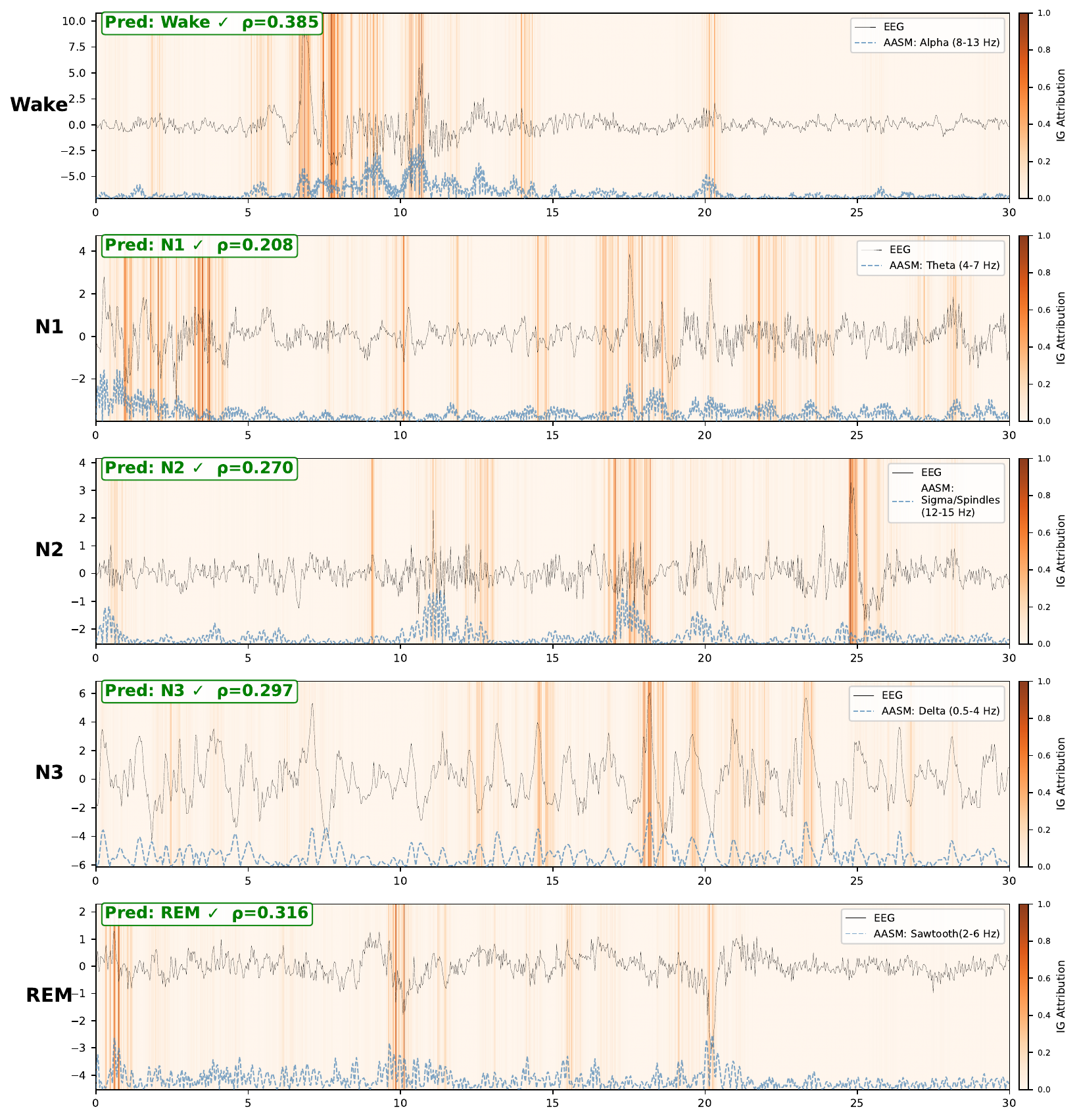}
  \caption{IG attribution heatmaps for the Normal OSA cohort 
  (AHI $<$ 5, F4-M1 channel). X-axis = time (seconds). 
  Y-axis = normalized EEG amplitude. Black line = raw EEG signal. 
  Orange = IG attribution intensity. Blue dashed = AASM-defined 
  frequency band envelope per stage.}
  \label{fig:ig_normal}
\end{figure}



Fig.~\ref{fig:ig_normal} shows IG attribution heatmaps for 
representative epochs from the Normal OSA cohort (AHI $<$ 5), which shows subjects with no OSA diagnosis 
exhibit the clearest AASM-defined EEG signatures, providing the most interpretable visual 
validation of attribution quality~\cite{berry2012aasm}. 
The orange heatmap directly visualizes the IG attribution 
scores computed on the CNN encoder output, while the blue 
dashed line shows the AASM frequency band envelope used as 
clinical ground truth for the Spearman correlation metric. 

Stage-specific observations confirm physiological alignment. 
In the Wake epoch, orange attribution concentrates 
between 5--12~s, coinciding with prominent alpha-band peaks 
visible in the blue envelope ($\rho=0.385$). 
In the N1 epoch, attribution is more diffuse but 
aligns with theta-band activity throughout the epoch, 
consistent with the mixed-frequency nature of light sleep. 
In the N2 epoch, attribution peaks near 5~s and 
25~s correspond to sleep spindle bursts clearly visible as 
localized blue peaks in the sigma band envelope ($\rho=0.270$). 
In the N3 epoch, orange attribution spans the full 
epoch and closely tracks delta wave peaks throughout, 
reflecting the sustained slow-wave activity characteristic 
of deep sleep consistent with N3 achieving the highest 
cohort-level IG correlations across all AHI groups (Tables~\ref{tab:xai_correlation}--\ref{tab:xai_correlation2}). 
In the REM epoch, attribution remains diffuse with 
no dominant peaks, consistent with the mixed-frequency 
low-amplitude EEG of REM sleep~\cite{berry2012aasm}. 
These stage-specific patterns confirm that IG correctly 
identifies physiologically meaningful time points rather 
than attributing importance to arbitrary signal 
artifacts~\cite{sujatha2023empirical}.

\begin{figure}[!ht]
  \centering
  \includegraphics[width=\columnwidth]{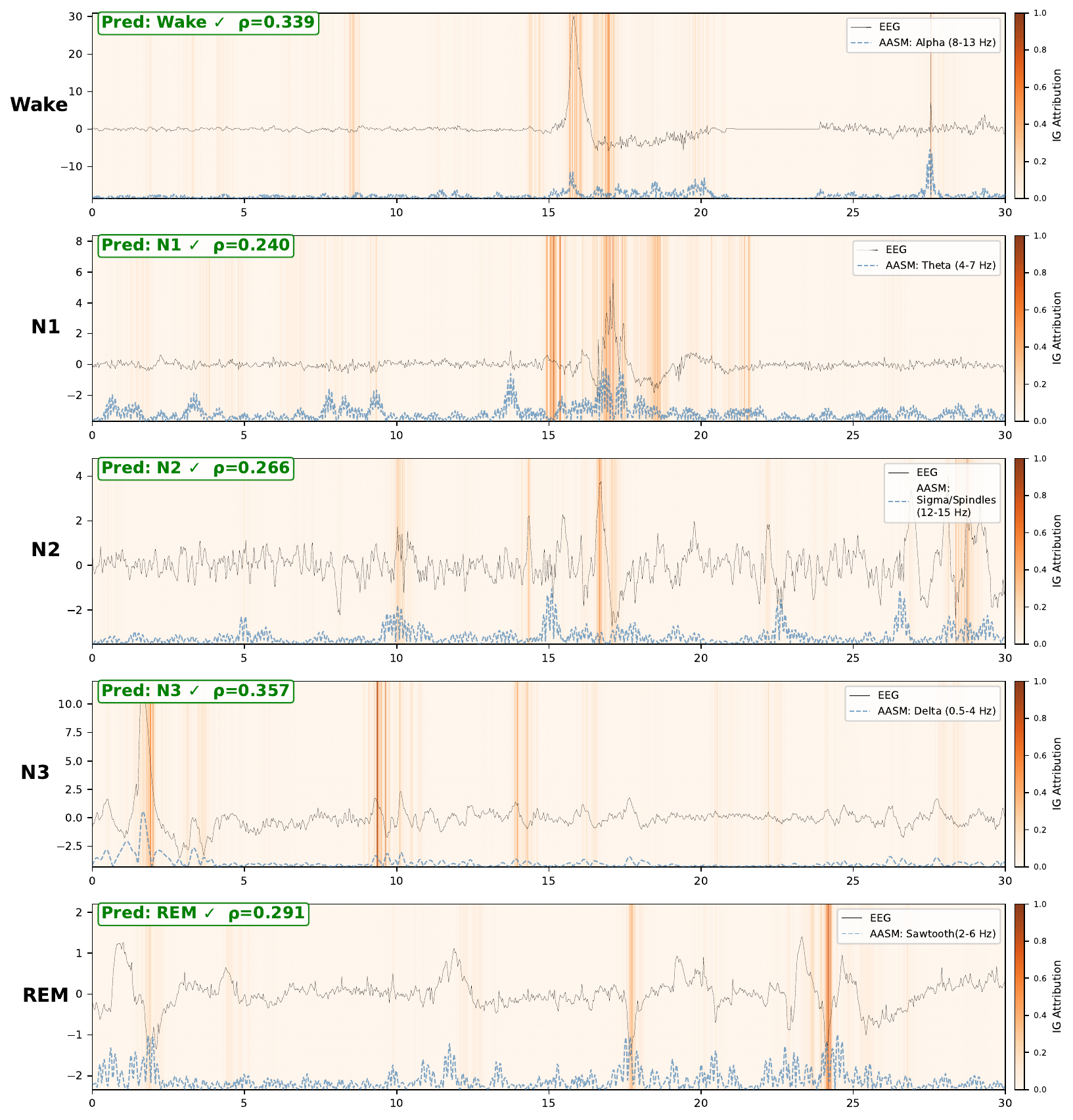}
  \caption{IG attribution heatmaps for the Severe OSA cohort 
  (AHI $\geq$ 30, F4-M1 channel). 
  Black line = raw EEG signal. Orange = IG attribution intensity. 
  Blue dashed = AASM-defined frequency band envelope per stage.}
  \label{fig:ig_severe}
\end{figure}

{Fig.~\ref{fig:ig_severe} shows IG attribution heatmaps for the Severe OSA cohort (AHI $\geq$ 30). 
Compared to the Normal cohort, attribution patterns become more 
diffuse and less aligned with the AASM band envelope, 
particularly for N3 where delta wave peaks are less pronounced 
due to OSA-induced slow-wave fragmentation. Cohort-level mean Spearman correlations for N3 decline from 0.314 (Normal) to 0.222 (Severe) for IG, as reported in Tables~\ref{tab:xai_correlation}--\ref{tab:xai_correlation2}; note that individual example epochs are selected as the highest-$\rho$ correctly classified instances and do not represent cohort averages. 
Mild and Moderate cohorts show intermediate attribution patterns, 
confirming a systematic relationship between OSA severity and 
attribution fidelity.} In Figs.~\ref{fig:ig_normal} and~\ref{fig:ig_severe}, epochs correspond to the highest-$\rho$ correctly classified examples per stage, selected to illustrate the clearest attribution patterns.

Overall, these results lead to three actionable conclusions for clinical XAI deployment. First, among the five evaluated methods, IG consistently achieves the highest physiological alignment across both healthy subjects and all four OSA severity groups, suggesting it may be a suitable choice for EEG-based sleep staging applications. Second, gradient-based methods without path integration (Grad-CAM, Vanilla Saliency) show weak alignment under this metric, producing near-zero correlations with AASM-defined frequency bands, particularly for N3 and REM. Third, the degradation of attribution quality with increasing 
OSA severity is not uniform across methods: IG maintains the 
highest mean correlation across all four AHI groups, while 
SHAP and DeepLIFT show greater variability across severity 
levels. These findings directly inform method selection for clinically interpretable sleep staging AI.

\section{Limitations}
\label{limitations}
This study has some limitations. First, the evaluation is restricted to gradient-based attribution methods. Perturbation-based approaches such as LIME and attention-based explanations were not evaluated and may offer complementary insights~\cite{arrieta2020explainable}. 
Second, class imbalance affects N1 and N3 stages across all AHI groups (Table~\ref{tab:ahi}), where these minority stages account for fewer than 15\% of epochs. N3 F1-scores are particularly low in OSA cohorts (approximately 33\% in the Normal DREAMT group); attribution maps for stages the model classifies poorly should be interpreted with caution, as low classification confidence may reduce the reliability of gradient-based explanations.
Third, attribution explanations are computed on the CNN encoder output rather than the final CRF prediction, as the CNN is the sole component responsible for feature learning from raw EEG. Developing CRF-aware attribution methods remains a direction for future work. Fourth, the XAI evaluation is conducted solely on the SleepXAI CNN-CRF architecture~\cite{dutt2023sleepxai}. Extending this analysis to other sleep staging architectures such as AttnSleep~\cite{eldele2024attnEEG} and transformer-based models would provide a broader understanding of attribution method behavior across different model families. Fifth, IG, DeepLIFT, and SHAP GradientExplainer multiply gradients by the input signal, so their attributions partly 
scale with input amplitude. Since the Spearman reference 
is also amplitude-based, some correlation may reflect this 
structural coupling. 


\section{Conclusion}
\label{sec:conclusion}

This paper presented the systematic evaluation of five gradient-based XAI attribution methods for single-channel EEG sleep staging using two datasets. We observed that Integrated Gradients consistently achieved the highest attribution fidelity compared to the other four XAI methods across all sleep stages and OSA severity groups (in the DREAMT dataset). We also confirmed this by measuring Spearman rank correlation with AASM-defined frequency bands. These findings highlight the importance of selecting appropriate XAI methods when developing clinically interpretable sleep staging systems. Future work will investigate perturbation-based and attention-based XAI methods to complement the gradient-based comparison presented here. We will also explore input-independent evaluation metrics, amplitude-normalized and trivial-attribution baselines, and model-randomization sanity checks to address potential confounding between gradient$\times$input attribution methods and amplitude-based reference signals.

\section*{Acknowledgment}
This work was supported in part by the NSF AI Institute for Foundations of Machine Learning (IFML).

\bibliographystyle{IEEEtran}
\bibliography{reference}

@article{dutt2023sleepxai,
  title={SleepXAI: An explainable deep learning approach for multi-class sleep stage identification},
  author={Dutt, Micheal and Redhu, Surender and Goodwin, Morten and Omlin, Christian W},
  journal={Applied Intelligence},
  volume={53},
  number={13},
  year={2023},
  publisher={Springer}
}

@misc{berry2012aasm,
  author    = {Berry, R. B. and Brooks, R. and Gamaldo, C. E. and Harding, S. M. and Marcus, C. and Vaughn, B. V.},
  title     = {The {AASM} Manual for the Scoring of Sleep and Associated Events},
  publisher = {American Academy of Sleep Medicine},
  year      = {2012}
}

@article{Younes_Raneri_Hanly_2016a,
  author = {Younes, M. and Raneri, J. and Hanly, P.},
  title = {Staging Sleep in Polysomnograms: Analysis of Inter-scorer Variability},
  journal = {Journal of Clinical Sleep Medicine},
  year = {2016}
}

@article{lajnef2015learning,
  author = {Lajnef, T. and others},
  title = {Learning Machines and Sleeping Brains},
  journal = {Journal of Neuroscience Methods},
  volume = {250},
  pages = {94--105},
  year = {2015}
}

@article{supratak2017deepsleepnet,
  author = {Supratak, A. and Dong, H. and Wu, C. and Guo, Y.},
  title = {{DeepSleepNet}: A Model for Automatic Sleep Stage Scoring},
  journal = {IEEE Trans. Neural Syst. Rehabil. Eng.},
  year = {2017}
}

@article{eldele2021attention,
  author = {Eldele, E. and others},
  title = {An Attention-Based Deep Learning Approach for Sleep Stage Classification},
  journal = {IEEE Trans. Neural Syst. Rehabil. Eng.},
  year = {2021}
}

@article{korkal2021sleepfragmentation,
  author = {Korkalainen, H. and others},
  title = {Detailed Assessment of Sleep Architecture with Deep Learning},
  journal = {IEEE J. Biomed. Health Inform.},
  volume = {25},
  number = {7},
  year = {2021}
}

@article{nassehi2024machine,
  author = {Nassehi, F. and others},
  title = {Machine Learning Based Severity Classification of Obstructive Sleep Apnea},
  journal = {Biomed. Signal Process. Control},
  year = {2024}
}

@misc{sharma2023transparency,
  author = {Sharma, S. and others},
  title = {Transparency in Sleep Staging},
  howpublished = {arXiv:2309.07156},
  year = {2023}
}

@article{sujatha2023empirical,
  author = {Ravindran, A. S. and Contreras-Vidal, J.},
  title = {An Empirical Comparison of Deep Learning Explainability Approaches for {EEG}},
  journal = {Scientific Reports},
  volume={13},
  number={1},
  pages={17709},
  year = {2023}
}

@article{samek2016evaluating,
  author = {Samek, W. and others},
  title = {Evaluating the Visualization of What a Deep Neural Network has Learned},
  journal = {IEEE Trans. Neural Netw. Learn. Syst.},
  year = {2016}
}

@article{lv2024ssleepnet,
  author = {Lv, X. and others},
  title = {{SSleepNet}: A Structured Sleep Network for Sleep Staging Based on Sleep Apnea Severity},
  journal = {Complex \& Intelligent Systems},
  year = {2024}
}

@inproceedings{sundararajan2017axiomatic,
  author = {Sundararajan, M. and Taly, A. and Yan, Q.},
  title = {Axiomatic Attribution for Deep Networks},
  booktitle = {ICML},
  pages = {3319--3328},
  year = {2017}
}

@inproceedings{shrikumar2017learning,
  author = {Shrikumar, A. and Greenside, P. and Kundaje, A.},
  title = {Learning Important Features Through Propagating Activation Differences},
  booktitle = {ICML},
  year = {2017}
}

@inproceedings{lundberg2017unified,
  author = {Lundberg, S. M. and Lee, S.-I.},
  title = {A Unified Approach to Interpreting Model Predictions},
  booktitle = {NeurIPS},
  volume = {30},
  year = {2017}
}

@misc{simonyan2013deep,
  author = {Simonyan, K. and Vedaldi, A. and Zisserman, A.},
  title = {Deep Inside Convolutional Networks: Visualising Image Classification Models and Saliency Maps},
  howpublished = {arXiv:1312.6034},
  year = {2013}
}

@inproceedings{selvaraju2017grad,
  author = {Selvaraju, R. R. and others},
  title = {{Grad-CAM}: Visual Explanations from Deep Networks via Gradient-based Localization},
  booktitle = {ICCV},
  pages = {618--626},
  year = {2017}
}

@article{zhao2022deep,
  author = {Zhao, D. and others},
  title = {A Deep Learning Algorithm Based on {1D CNN-LSTM} for Automatic Sleep Staging},
  journal = {Technology and Health Care},
  volume = {30},
  number = {2},
  pages = {323--336},
  year = {2022}
}

@inproceedings{glorot2010understanding,
  author = {Glorot, X. and Bengio, Y.},
  title = {Understanding the Difficulty of Training Deep Feedforward Neural Networks},
  booktitle = {AISTATS},
  pages = {249--256},
  year = {2010}
}

@inproceedings{lafferty2001conditional,
  author = {Lafferty, J. and McCallum, A. and Pereira, F. C.},
  title = {Conditional Random Fields: Probabilistic Models for Segmenting and Labeling Sequence Data},
  booktitle = {ICML},
  pages = {282--289},
  year = {2001}
}

@article{kemp2000sleepedf,
  author = {Kemp, B. and others},
  title = {Analysis of a Sleep-dependent Neuronal Feedback Loop},
  journal = {IEEE Trans. Biomed. Eng.},
  volume = {47},
  number = {9},
  pages = {1185--1194},
  year = {2000}
}

@misc{wang2025dreamt,
  author = {Wang, K. and Yang, J. and Shetty, A. and Dunn, J.},
  title = {{DREAMT}: Dataset for Real-time Sleep Stage Estimation Using Multisensor Wearable Technology},
  howpublished = {PhysioNet},
  year = {2025}
}

@article{lauritsen2020explainable,
  author = {Lauritsen, S. M. and others},
  title = {Explainable Artificial Intelligence Model to Predict Acute Critical Illness from {EHR}},
  journal = {Nature Communications},
  year = {2020}
}

@article{arrieta2020explainable,
  author = {Arrieta, A. B. and others},
  title = {Explainable Artificial Intelligence ({XAI}): Concepts, Taxonomies, Opportunities and Challenges toward Responsible {AI}},
  journal = {Information Fusion},
  volume = {58},
  pages = {82--115},
  year = {2020}
}

@inproceedings{supratak2020tinysleepnet,
  author    = {Supratak, A. and Guo, Y.},
  title     = {{TinySleepNet}: An Efficient Deep Learning Model for Sleep 
               Stage Scoring based on Raw Single-Channel {EEG}},
  booktitle = {42nd Annual International Conference of the IEEE EMBC},
  pages     = {641--644},
  year      = {2020}
}

@article{back2019iitnet,
  author    = {Back, S. and Lee, S. and Seo, H. and Park, D. and 
               Kim, T. and Lee, K.},
  title     = {Intra- and Inter-epoch Temporal Context Network ({IITNet}) 
               Using Sub-epoch Features for Automatic Sleep Scoring on 
               Raw Single-Channel {EEG}},
  journal   = {Biomedical Signal Processing and Control},
  volume    = {60},
  pages     = {102037},
  year      = {2020}
}

@article{phan2022xsleepnet,
  author    = {Phan, H. and Ch{\'e}n, O. Y. and Koch, P. and 
               Mertins, A. and De Vos, M.},
  title     = {{XSleepNet}: Multi-view Sequential Model for Automatic 
               Sleep Staging},
  journal   = {IEEE Transactions on Pattern Analysis and Machine Intelligence},
  year      = {2022}
}

@article{muller2021trustworthy,
  author = {M{\"u}ller, V. C.},
  title = {Ethics of Artificial Intelligence and Robotics},
  journal = {Stanford Encyclopedia of Philosophy},
  year = {2021}
}

@article{topol2019high,
  author = {Topol, E. J.},
  title = {High-Performance Medicine: The Convergence of Human and Artificial Intelligence},
  journal = {Nature Medicine},
  volume = {25},
  pages = {44--56},
  year = {2019}
}

@article{tallal2026modulation,
  title={Modulation-Based Feature Extraction for Robust Sleep Stage Classification Across Apnea-Based Cohorts},
  author={Tallal, Unaza and Agrawal, Rupesh and Kshirsagar, Shruti},
  journal={Biosensors},
  volume={16},
  number={1},
  pages={56},
  year={2026},
  publisher={MDPI}
}

@book{kshirsagar2022affective,
  title={Affective Human-Machine Interfaces: Towards Multi-lingual, Environment-Robust Emotion Detection from Speech},
  author={Kshirsagar, Shruti Rajendra},
  year={2022},
  publisher={Institut National de la Recherche Scientifique (Canada)}
}

@article{Sousa2023,
  author={Sousa, C. and others},
  title={Explainability in deep learning for EEG sleep stage classification},
  journal={Computers in Biology and Medicine},
  year={2023},
  volume= {155},
  pages   = {106651},
  doi     = {10.1016/j.compbiomed.2023.106651}
}

@inproceedings{ancona2018towards,
  title={Towards better understanding of gradient-based attribution methods for deep neural networks},
  author={Ancona, Marco and Ceolini, Enea and {\"O}ztireli, Cengiz and Gross, Markus},
  booktitle={International Conference on Learning Representations (ICLR)},
  year={2018}
}

@article{kshirsagartrustworthy,
  title={Trustworthy AI for Safety-Critical Perception and Decision Systems},
  author={Kshirsagar, Shruti},
  journal={IEEE SMC Feature article},
  year={2026}
}

@article{hossain2026demographic,
  title={Demographic-Aware Transfer Learning for Sleep Stage Classification in Clinical Polysomnography},
  author={Hossain, SM and Kshirsagar, Shruti},
  journal={arXiv preprint arXiv:2605.02245},
  year={2026}
}

@article{tallal2026stda,
  title={STDA-Net: Spectrogram-Based Domain Adaptation for cross-dataset Sleep Stage Classification},
  author={Tallal, Unaza and Kshirsagar, Shruti and Shukla, Ankita},
  journal={arXiv e-prints},
  pages={arXiv--2605},
  year={2026}
}

@article{eldele2024attnEEG,
  title={Attention-Based Transformer Models for EEG-Based Sleep Stage Classification: A Systematic Review},
  author={Eldele, Emadeldeen and Chen, Zhenghua and Liu, Chengyu},
  journal={Biomedical Signal Processing and Control},
  volume={89},
  pages={105--118},
  year={2024}
}

@article{chen2024xaisleep,
  title={Explainability in Deep Learning-Based Sleep Stage Classification: Challenges and Opportunities},
  author={Chen, Jing and Wang, Li and Liu, Yang},
  journal={Computers in Biology and Medicine},
  volume={170},
  pages={107--921},
  year={2024}
}

\end{document}